\documentclass[sigconf,review]{llncs} 

\usepackage{graphicx} 
\usepackage{makecell}
\usepackage{tabularx}
\usepackage{url}
\usepackage{hyperref}
\title{Are Gradient Boosting Models Suitable for Intermittent Demand Forecasting?}

\author{Vladislav Kislinskii \and Mazhar Hameed}

\institute{
Gisma University of Applied Sciences, Potsdam, Germany \\
\email{Vladislav.Kislinskii@gisma-student.com, mazhar.hameed@gisma.com}
}

\begin{document}

\maketitle

\begin{abstract}
Demand forecasting is critical in modern industry, offering opportunities to reduce costs and gain competitive advantage through improved inventory management. However, forecasting becomes particularly challenging for products with intermittent demand, where demand occurs infrequently and time series contain many zero observations. Such dynamics are common across diverse sectors, such as industrial organizations, consumer goods, aviation, automotive, and electronics. Motivated by these challenges, this paper explores the potential of gradient boosting models to improve forecasting performance. We evaluate statistical, specialized, machine learning, and ensemble approaches across multiple datasets. The results show that specialized methods achieve the strongest performance among individual models, while gradient boosting on its own tends to underperform. However, combining a machine learning model with a specialized approach improves forecasting accuracy by up to 10\%, demonstrating that even simple ensembles can outperform single models. Overall, the findings highlight the value of combining machine learning with domain-specific forecasting techniques for intermittent demand.

\end{abstract}

\keywords{Intermittent Demand Forecasting, Gradient Boosting, Ensemble Methods, Time Series Forecasting}

\section{Intermittent Demand Forecasting}
\label{sec:intro}

Demand forecasting enables companies to capture complex historical patterns and predict future demand, supporting more effective management of inventory, supply chains, and operations. However, not all demand patterns are easily predictable. Some products are sold rarely
and inconsistently, a phenomenon known as \emph{intermittent demand}~\cite{kostenko2006note}~\cite{syntetos2005categorization}.  It is
characterized by sporadic and irregular sales dynamics, with a high proportion of zero-demand periods. Typical examples include spare parts, seasonal, luxury, and specialized products for solving rare problems. In industrial organizations, 60\% of the stock-keeping units (SKUs) have intermittent demand dynamics ~\cite{johnston2003examination}. More broadly, all slow-moving items fall into this category, regardless of whether a company sells them directly or uses them internally in its operations.

The demand dynamics of such products present a significant forecasting challenge, and many methods perform poorly when applied to this type of data. Inaccurate predictions have serious consequences: overstock ties up capital and increases holding costs, while understock leads to shortages and lost sales. Therefore, developing effective systems for intermittent demand forecasting can deliver substantial benefits to companies.

Given the importance and complexity of this problem, extensive research has been carried out in this area. Specialized approaches to address it can generally be classified into five categories:

\begin{itemize}
    \item Classical time series approaches, such as SES~\cite{brown1956exponential} and ETS~\cite{hyndman2002state}, were originally designed for continuous data but have become widely adopted in both academia and industry across a variety of demand patterns.

    \item Parametric approaches (Croston’s~\cite{croston1972forecasting}, SBA~\cite{syntetos2005accuracy}, TSB~\cite{teunter2011intermittent}) --- the first group of specialized techniques for intermittent demand, which assume specific demand characteristics and are relatively inflexible in their calculations.

    \item Nonparametric aggregation frameworks (ADIDA~\cite{nikolopoulos2011adida},  IMAPA~\cite{petropoulos2015forecast})  --- algorithms that aggregate the original data to mitigate the impact of multiple zero values.  

    \item Machine learning models (XGBoost~\cite{chen2016xgboost}, CatBoost~\cite{prokhorenkova2018catboost}) --- methods not originally developed for intermittent time series forecasting, but capable of leveraging multiple features in contrast to traditional univariate time series approaches.  

    \item Ensembles (CAD~\cite{song2024aggregate}, FIDE~\cite{li2022feature}, DIVIDE~\cite{li2022feature}, THieF~\cite{kourentzes2021elucidate}) --- systems that combine multiple algorithms to produce more accurate forecasts.
\end{itemize}

Despite these advances, several gaps remain. First, there is a lack of comparative studies that directly evaluate traditional time series methods against gradient boosting algorithms. Second, few studies explore hybrid approaches that combine specialized intermittent demand techniques with gradient boosting models. 
To address these gaps, this study conducts an experimental evaluation of different approaches to intermittent demand forecasting: classical (ETS), specialized (SBA, TBS, ADIDA, IMAPA), machine learning (XGBoost, CatBoost), and hybrid methods (specialized + machine learning). 

The main objectives are:
\begin{itemize}
    \item To evaluate how classical and specialized methods perform in comparison to gradient boosting models.
    \item To investigate whether gradient boosting can enhance the performance of specialized methods.
    \item To examine how model performance varies across different evaluation metrics, and whether these differences reveal additional insights.
\end{itemize}

With this study, we aim to provide a more comprehensive understanding of existing methods and thereby contribute to the advancement of this field. By addressing aspects not previously explored in the literature, our work offers value to both the academic community and practitioners seeking to improve forecasting in organizational contexts. All artifacts are publicly available at GitHub\footnote{\url{https://github.com/vkislinskii/intermittent_demand_forecasting_paper}}.

The rest of the paper is structured as follows: Section~\ref{sec:Environment_setup} describes the experimental setup and key implementation details. Section~\ref{sec:evaluation} presents the results and discusses their implications. Finally, Section~\ref{sec:Conclusion} summarizes the study and outlines directions for future research.

\section{Environment and Setup}
\label{sec:Environment_setup}

The models’ predictions were evaluated using a unified experimental setup. A total of eight algorithms were assessed on the Royal Air Force (RAF) dataset~\cite{teunter2009forecasting}, including a classical method (ETS), specialized approaches (SBA, TSB, ADIDA, IMAPA), machine learning models (XGBoost, CatBoost), and a hybrid system (IMAPA + CatBoost). All implementations are available in Python: the time series algorithms through the \texttt{statsforecast} package, and CatBoost and XGBoost via their respective libraries.

The \texttt{statsforecast} implementations do not support cross- validation for forecasting future values, but only for assessing performance on the training data. Consequently, all time series models (SBA, TSB, ADIDA, IMAPA, ETS) were evaluated using a simple train–test split. For each prediction horizon, performance was measured on the test subset (the last period of the series), while the preceding data was used for model training.

CatBoost and XGBoost models, in contrast, were trained using the train-validation-test split. The validation set was used for feature selection and hyperparameter tuning. Before the final performance evaluation on the test subset, the model with the optimal features and hyperparameters was retrained on the combined training and validation data.

A common approach is to set the validation part the same size as the test part. However, this strategy may lead to a poor performance when the validation set constitutes only a small fraction of the entire dataset. In such cases, the validation data can differ substantially from the test subset, causing the model to optimize the target metric incorrectly and resulting in poor generalization.

This problem is especially important for the intermittent time series data, where scarcity of demand occurrences is the primary challenge. The RAF dataset has 2 forecast horizons (3 and 6), which represent a small number of observations, making this problem especially relevant (see~Table~\ref{tab:horizons_dataset_share}). To address this and ensure robust model training, the validation period for the CatBoost and XGBoost models was set to 12 months.

\begin{table}[h!]
    \centering
    \renewcommand{\arraystretch}{1.3} 
    \begin{tabular}{c|c}
        \textbf{Forecast horizon (months)} & \textbf{Dataset share} \\
        \hline\hline
        3 & 3.93\% \\
        \hline
        6 & 7.86\% \\
        \hline
        12 & 15.7\% \\
    \end{tabular}
    \caption{Forecast horizons' dataset share}
    \label{tab:horizons_dataset_share}
\end{table}

Accordingly, for the ensemble, this nuance was taken into account: both the demand quantity and demand occurrence prediction models employed a train-validation-test split, following the sequence below:
\begin{itemize}
    \item The regression model was trained on the combined training and validation subset.
    \item The classification model used the training and validation data to identify the most important features and tune hyperparameters. After this, the final model was retrained on the combined training and validation data.
    \item The outputs of both models were combined, and the final prediction was evaluated on the test subset.
\end{itemize}

\section{Performance Evaluation}
\label{sec:evaluation}

In this section, we first describe the datasets used in the study. We then present the evaluation metrics and experimental setup, and finally report the results.

\subsection{Data}
\label{ssec:data}
\subsubsection{High-level data description}

For this study, the Royal Air Force (RAF) dataset was used. Introduced by Teunter and Duncan~\cite{teunter2009forecasting}, it contains monthly demand data for 5,000 spare parts used by the United Kingdom’s air and space force over the period from January 1996 to December 2002.

\begin{figure}[h]
    \centering
    \includegraphics[width=0.5\textwidth]{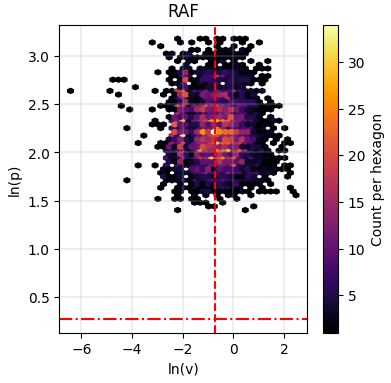}
    \caption{Natural logarithms of $p$ and $v$ distribution}
    \label{fig:demand-hex} 
\end{figure}

Figure~\ref{fig:demand-hex} presents a density hexagon plot showing the distribution of time series across the natural logarithm of the inter-demand interval ($p$) and the squared coefficient of variation ($v$), with color intensity representing the number of cases within each area. The observations in the dataset are relatively homogeneous, as most values are concentrated within a narrow range.

\subsubsection{Feature engineering}
All the considered time series models (SBA, TSB, ADIDA, IMAPA, ETS) are univariate approaches, and therefore do not incorporate additional features. In contrast, CatBoost and XGBoost models can theoretically benefit from supplementary information. For this purpose, several feature groups were introduced, including calendar variables, lag features, rolling means, and trend-related attributes (see Table~\ref{tab:raf-add-attributes}).

\begin{table}[h]
    \centering
    \begin{tabular}{l|l|l}
        \textbf{Feature} & \textbf{Description} & \textbf{Example} \\
        \hline\hline
        ds\_start & First sale month & 1996-01-01 \\
        \hline
        month & Month & 4 \\
        \hline
        quarter & Quarter & 2 \\
         \hline
        year & Year & 1996 \\
         \hline
        qty\_lag1 & Demand last month & 0 \\
         \hline
        qty\_lag2 & Demand two months ago & 2 \\
         \hline
        qty\_lag3 & Demand three months ago & 6 \\
         \hline
        bin\_lag1 & \makecell[l]{Binary flag of demand \\ occurrence last month} & 0 \\
         \hline
        bin\_lag2 & \makecell[l]{Binary flag of demand \\ occurrence two months ago} & 1 \\
         \hline
        bin\_lag3 & \makecell[l]{Binary flag of demand \\ occurrence three months ago} & 1 \\
         \hline
        qty\_roll\_mean\_3 & \makecell[l]{Average demand over the \\ last three months} & 2.667 \\
         \hline
        bin\_roll\_mean\_3 & \makecell[l]{Average binary flag of demand \\ occurrence over the last three \\ months} & 0.667 \\
         \hline
        qty\_roll\_mean\_6 & \makecell[l]{Average demand over \\ the last six months}  & 0 \\
         \hline
        qty\_diff1 & \makecell[l]{The difference between last \\ month and the month \\ before} & -2 \\
         \hline
        qty\_trend\_ratio & \makecell[l]{Relation of the last 3-month \\ average vs the last 6-month \\ average} & 2,666,667 \\
    \end{tabular}
    \caption{RAF dataset additional features}
    \label{tab:raf-add-attributes}
\end{table}

\subsection{Evaluation Metrics}
\label{ssec:eval_metr}
Two metrics were employed to measure demand forecast quality. The Root Mean Squared Scaled Error (RMSSE) provides a high-level measure of prediction accuracy across series, while the Mean Absolute Error (MAE) reports the error in the scale of the target variable, making it straightforward to interpret. Together, these metrics offer a balanced evaluation framework.

For evaluating the classifier model used in the ensemble, the F1-score was applied, as it balances precision (correctness) and recall (sensitivity) to provide a robust measure of classification performance.

\subsection{Experiments}
\label{ssec:exp}
For evaluation, forecasting horizons of 3, 6, and 12 months were considered~\cite{li2022feature} 

\subsubsection{Performance of single models: ranking and comparison}

By default, all the prediction models return a prediction in decimal values. Assessing performance solely on this basis is unrealistic, since retail goods are not sold in fractional quantities.
As a result, metrics differ depending on whether predictions are rounded before evaluation. To account for this, model performance is reported in both forms (see Table~\ref{tab:forecast-raf}). Evaluating models without rounding enables a more accurate comparison across methods, while evaluating with rounding reflects conditions closer to real-world business environments.

\begin{table}[h]
    \centering
    \begin{tabularx}{0.73\linewidth}{l|l|l|l|l|l}
   \textbf{\makecell[l]{Forecast \\horizon \\(months)}} & \textbf{Method} & \textbf{\makecell[l]{RMSSE \\(float)}} & \textbf{\makecell[l]{RMSSE \\(integer)}} & \textbf{\makecell[l]{MAE \\(float)}} & \textbf{\makecell[l]{MAE \\(integer)}}\\
    \hline\hline
    3 & SBA & 0.512 & 0.434 & 2.767 & 2.693 \\
    \hline
    & TSB & 0.456 & 0.409 & 2.379 & 2.321 \\
    \hline
    & ADIDA & 0.466 & 0.4 & 2.419 & 2.339 \\
    \hline
    & IMAPA & 0.462 & 0.399 & 2.4 & 2.323 \\
    \hline
    & ETS & 0.482 & 0.413 & 2.573 & 2.498 \\
    \hline
    & CatBoost & 0.71 & 0.674 & 2.315 & 2.279 \\
    \hline
    & XGBoost & 0.952 & 1.032 & 2.557 & 2.623 \\
    \hline  \hline
    6 & SBA & 0.575 & 0.517 & 2.727 & 2.654 \\
    \hline
    & TSB & 0.537 & 0.5 & 2.26 & 2.198 \\
    \hline
    & ADIDA & 0.538 & 0.489 & 2.337 & 2.261 \\
    \hline
    & IMAPA & 0.536 & 0.488 & 2.316 & 2.241 \\
    \hline
    & ETS & 0.553 & 0.502 & 2.51 & 2.438 \\
    \hline
    & CatBoost & 0.808 & 0.788 & 2.261 & 2.218 \\
    \hline
    & XGBoost & 1.03 & 1.072 & 2.509 & 2.542 \\
    \hline \hline
    12 & SBA & 0.626 & 0.595 & 2.666 & 2.592 \\
    \hline
    & TSB & 0.624 & 0.597 & 2.283 & 2.22 \\
    \hline
    & ADIDA & 0.601 & 0.575 & 2.305 & 2.228 \\
    \hline
    & IMAPA & 0.604 & 0.578 & 2.285 & 2.215 \\
    \hline
    & ETS & 0.617 & 0.59 & 2.461 & 2.387 \\
    \hline
    & CatBoost & 0.704 & 0.703 & 1.93 & 1.906 \\
    \hline
    & XGBoost & 1.268 & 1.101 & 2.533 & 2.356 \\
    \end{tabularx}
    \caption{Metrics for RAF dataset forecast} 
    \label{tab:forecast-raf} 
\end{table}

Several key observations emerge from the results:
\begin{enumerate}
    \item RMSSE and MAE rankings differ substantially for CatBoost across all forecasting horizons of the RAF dataset. The model performs among the best according to MAE but among the worst according to RMSSE. This discrepancy arises from the metrics’ different error sensitivities: while CatBoost generally avoided large average errors (yielding a small MAE), it occasionally produced significant errors that inflated RMSSE.

    \item ADIDA, in its standard implementation, faces the challenge of selecting an appropriate aggregation level for specific dataset. IMAPA addresses this by using multiple aggregation levels simultaneously, which should in principle yield better performance. However, in our experiments the two methods produced nearly identical results (differences below 1\%), because in the \texttt{statsforecast} package ADIDA automatically selects the optimal aggregation level, equal to the mean inter-demand interval~\cite{nixtlaADIDA}.

    \item The relative rankings across models can be summarized as follows:  
    \begin{itemize}  
    \item XGBoost consistently delivered the weakest results across both metrics.  
    \item CatBoost performed poorly according to RMSSE but ranked highest under MAE.  
    \item TSB, IMAPA, and ADIDA showed strong and consistent accuracy across all evaluation settings.  
    \item ETS achieved moderate performance.  
    \item SBA performed worst under MAE, but mid-range under RMSSE, resulting in an overall mediocre ranking.  
    \end{itemize}  

\end{enumerate}

\subsubsection{Ensemble performance}
To investigate the impact on the prediction accuracy, the ensemble was designed as follows:

\begin{itemize}
    \item Demand quantity prediction --- IMAPA (selected as the best-performing specialized model).
    \item Demand occurrence prediction --- CatBoost (chosen over XGBoost due to superior performance).
\end{itemize}

Since ensemble accuracy depends heavily on the CatBoost configuration, a structured process was applied:

\begin{itemize}
    \item A baseline model was trained using the initial feature set.
    \item Additional features were introduced, and the most impactful ones were retained for training. 
    \item The model’s hyperparameters were tuned to further improve performance.
\end{itemize}

The F1-score was computed on the validation subset iteratively for each setup, and any step that resulted in a downgrade was skipped. Table~\ref{tab:forecast-raf-ens} presents the assessment of the ensemble approach on the test subset. Across all planning horizons, incorporating CatBoost improved the target metrics. However, this improvement may not be entirely sustainable, as the classifier model faces challenges in reliably defining class boundaries. Consequently, the general applicability of the ensemble depends strongly on both the dataset and the prediction horizon. The intermediate F1-scores are omitted, as they were only used for model selection and are not relevant to the final evaluation based on RMSSE and MAE.

\begin{table}[h]
    \centering
    \renewcommand{\arraystretch}{1.3} 
    \begin{tabular}{c|l|l|l|l|l}
        \textbf{\makecell[l]{Forecast \\horizon \\(months)}} & \textbf{Method} & \textbf{\makecell[l]{RMSSE \\(float)}} & \textbf{\makecell[l]{RMSSE \\(integer)}} & \textbf{\makecell[l]{MAE \\(float)}} & \textbf{\makecell[l]{MAE \\(integer)}}\\
        \hline \hline
        3 & IMAPA & 0.462 & 0.399 & 2.4 & 2.323 \\   
        \hline
         & \makecell[l]{IMAPA + \\CatBoost} & 0.417 & 0.376 & 1.802 & 1.762 \\
         \hline
         & \textit{difference} & 9.74\% & 5.76\% & 24.92\% & 24.15\% \\
        \hline  \hline
         6 & IMAPA & 0.536 & 0.488 & 2.316 & 2.241 \\
         \hline
         & \makecell[l]{IMAPA + \\CatBoost} & 0.520 & 0.478 & 1.879 & 1.825 \\
         \hline
         & \textit{difference} &  2.99\% & 2.05\% & 18.87\% & 18.56\% \\
         \hline  \hline
        12 & IMAPA & 0.604 & 0.578 & 2.285 & 2.215 \\
        \hline
         & \makecell[l]{IMAPA + \\CatBoost} & 0.596 & 0.572 & 1.852 & 1.808 \\
         \hline
         & \textit{difference} &  1.32\% & 1.04\% & 18.95\% & 18.37\% \\
    \end{tabular}
    \caption{Ensemble forecast metrics for RAF dataset}
    \label{tab:forecast-raf-ens}
\end{table}
\section{Conclusion}
\label{sec:Conclusion}

In this paper, we evaluated a range of approaches to intermittent demand forecasting, including classical time series methods, specialized algorithms, gradient boosting models, and hybrid techniques. Using the RAF dataset, the study compared their performance across multiple evaluation metrics.

The results showed that model rankings varied considerably depending on the chosen metric and the use of rounding, with differences averaging around 5\% and in some cases reaching up to 18\%. Among individual models, IMAPA achieved the strongest results, while CatBoost performed best under the MAE metric but poorly under RMSSE, and XGBoost consistently ranked lowest. CatBoost also exhibited less predictable behavior: unlike time series models, whose accuracy typically declined with longer horizons, CatBoost maintained similar performance across both short and long horizons. Importantly, the hybrid approach combining IMAPA with CatBoost improved accuracy by up to 10\% over IMAPA alone.

Overall, this paper highlights the potential of integrating machine learning with specialized intermittent demand forecasting techniques. While the proposed combination is not a universal solution, it provides a simple and effective way to improve forecast accuracy and contributes to advancing this area of research.

\bibliography{bibliography}

\end{document}